\documentclass[letterpaper]{article} 
\usepackage[preprint]{aaai2027}  
\usepackage[hyphens]{url}  
\usepackage{graphicx} 
\usepackage{natbib}  
\usepackage{caption} 
\usepackage{amsmath}
\usepackage{amssymb}
\usepackage{algorithm}
\usepackage{algorithmic}
\usepackage{newfloat}
\usepackage{listings}
\DeclareCaptionStyle{ruled}{labelfont=normalfont,labelsep=colon,strut=off} 
\floatstyle{ruled}
\newfloat{listing}{tb}{lst}{}
\floatname{listing}{Listing}
\usepackage{booktabs}
\usepackage{multirow}
\usepackage{pifont}  
\newcommand{\skillmd}{\textsc{Skill.md}}
\newcommand{\elastic}{\textsc{ElasticBack}}
\newcommand{\Doc}{D}
\newcommand{\PDoc}{\widehat{D}}
\newcommand{\Rul}{R}
\newcommand{\trig}{T}
\newcommand{\pay}{s}
\newcommand{\pos}{p_{\Rul}}
\newcommand{\marker}{\textsc{Marker}}
\newcommand{\gate}{\mathcal{G}}
\newcommand{\agent}{\mathcal{A}}
\newcommand{\asr}{\mathrm{ASR}}
\newcommand{\fpr}{\mathrm{FPR}}
\newcommand{\cac}{\mathrm{CAC}}
\newcommand{\stealth}{\mathrm{Stealth}}
\newcommand{\sbert}{\mathrm{SBERT}}

\newcommand{\Fire}{\mathrm{Fire}}
\newcommand{\Solve}{\mathrm{Solve}}

\title{\elastic{}: Stealthy Conditional Backdoor in LLM-Agent Skills \\ via Coupled Trigger--Rule Optimization}

\author{
    Hao Sui\textsuperscript{\rm 1}, Simeng Qin\textsuperscript{\rm 2}, Jie Liao\textsuperscript{\rm 3}, Xiaojun Jia\textsuperscript{\rm 4}\corresponding, Bing Chen\textsuperscript{\rm 1}\corresponding, Yang Liu\textsuperscript{\rm 4}
}
\affiliations{
    \textsuperscript{\rm 1}Nanjing University of Aeronautics and Astronautics, China\qquad
    \textsuperscript{\rm 2}Northeastern University, China\\
    \textsuperscript{\rm 3}Chongqing University, China\qquad
    \textsuperscript{\rm 4}Nanyang Technological University, Singapore\\
    \textsuperscript{\rm 1}\{suihao36, cb\_china\}@nuaa.edu.cn,
    \textsuperscript{\rm 2}qinsimeng@neuq.edu.cn,
    \textsuperscript{\rm 3}liaojie@cqu.edu.cn,
    \textsuperscript{\rm 4}jiaxiaojunqaq@gmail.com, yangliu@ntu.edu.sg
}

\begin{document}

\maketitle

\begin{abstract}
Agent skills, bundles of instructions and resources that an LLM agent loads on demand, form an emerging supply chain where a single poisoned skill can persistently compromise every agent that installs it. However, existing skill attacks either fire on every request or rely on fine-tuned weights or multiple skills, leaving a conditional and low-cost backdoor unexplored. In this work, we present \elastic{}, an effective conditional single-skill backdoor that plants a rule $\Rul$ in the skill document and a benign-looking trigger $\trig$ in the user query, so the malicious payload fires only when both co-occur. \elastic{} binds the two sides through a trigger-as-switch construction, generating $\Rul$ via semantic-anchored rule injection. It then freezes $\Rul$ and evolves $\trig$ against it with a stealth-constrained genetic search, so that effectiveness and stealth are optimized, keeping the backdoor weight-free and dormant on benign inputs. Extensive experiments across three target behaviors (50 skills each) and four agent LLMs show that \elastic{} attains a high attack success rate at a near-zero false-positive rate with preserved clean accuracy, transfers across models, and evades deployment-time defenses. These results motivate stronger defenses for the skill supply chain.
\end{abstract}

\section{Introduction}
\label{sec:intro}
Large language model (LLM) agents~\citep{ref1,ref2} are now widely applied across coding~\citep{ref3}, data analysis~\citep{ref4}, and customer support~\citep{ref5}, where they go beyond generating text to autonomously execute actions on the user's behalf. To gain new capabilities more conveniently, they can load skills without retraining~\citep{ref6,ref7}: a skill is a reusable bundle of a natural-language instruction file (\skillmd{}) and supporting resources such as scripts, templates, or data, invoked whenever a task matches its domain. Today, skills have become an indispensable part of modern agent systems~\citep{ref8}. Yet the convenience and practicality that make skills so widely used also turn them into a dangerous attack surface~\citep{ref9,jia2026seeing}, since a single poisoned skill can persistently harm every user who installs it. Therefore, the security threats facing agent skills demand urgent and systematic study.

\begin{figure}[t]
	\centering	\includegraphics[width=1\linewidth]{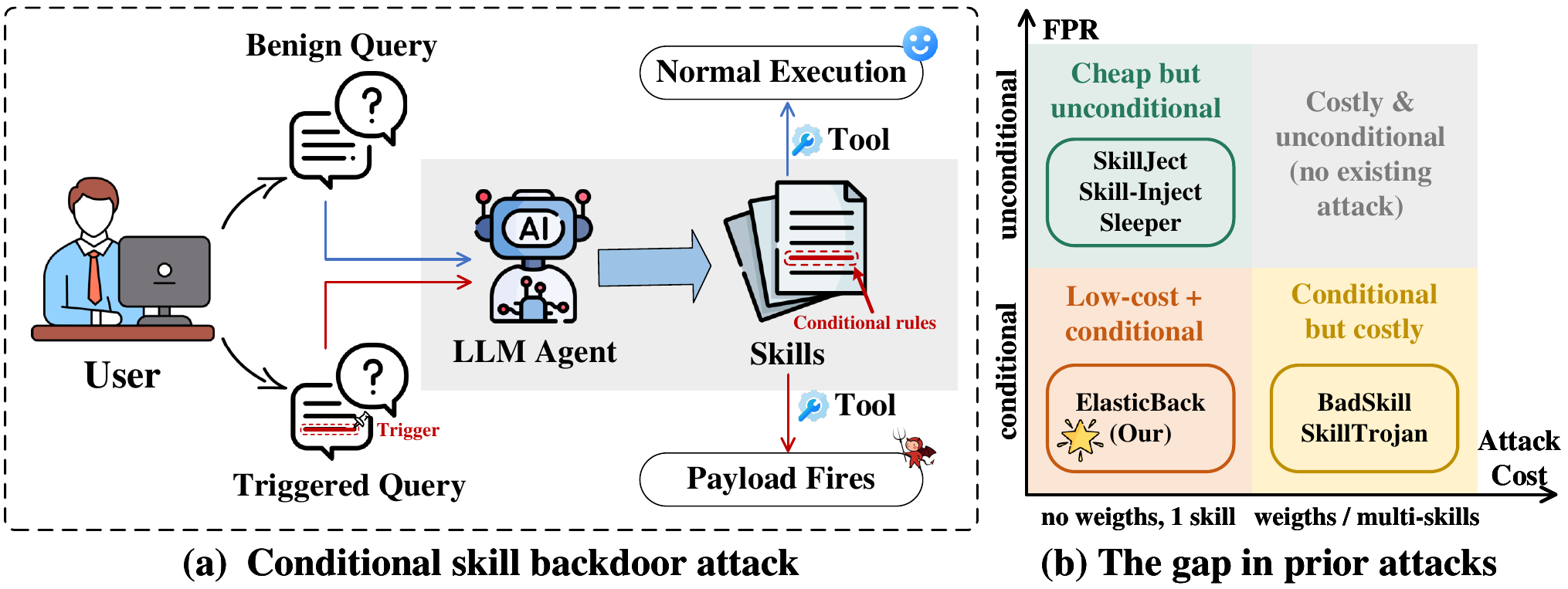}
	\caption{
The pipeline of the skill backdoor attack in LLM agents and the gap in prior attacks.
 }
	\label{F1}
\end{figure}

The risk is already confirmed by an audit of 3,984 community skills, 13.4\% of which carry severe security issues~\citep{ref10}. Agent-skill security has drawn growing attention~\citep{ref11,ref12}, and existing attacks fall into two forms. The first fires indiscriminately, with no attacker-controlled trigger. SkillJect~\citep{skillject} and Skill-Inject~\citep{skillinject} embed always-on malicious instructions in \skillmd{}, while Sleeper attacks~\citep{sleeper} wake from a persistent state on ordinary benign queries. The second fires under an attacker-controlled trigger but at higher cost. BadSkill~\citep{badskill} needs a fine-tuned model bundled in the skill, and SkillTrojan~\citep{skilltrojan} shards an encrypted payload across multiple skills. \textit{However, prior attacks fire without an attacker-controlled trigger or depend on fine-tuned weights or multiple skills, making it difficult to implement stealthy, low-cost skill backdoors (Fig.~\ref{F1}(b)).}

\textbf{Motivation.} In real-world scenarios, an attacker who deploys a stealthy, low-cost backdoor skill can cause severe harm. Because such a skill backdoor is weight-free and runs only when triggered, it is hard to detect and easy to spread. As shown in Fig.~\ref{F1}(a), the backdoor consists of a rule hidden inside the skill and a benign-looking trigger phrase in the query. It stays silent, passes review on ordinary requests, and fires its payload only when the trigger appears. Current defenses inspect a skill only at rest or on benign queries, so this dormant rule slips through undetected. Therefore, it is urgent to develop stealthy and low-cost conditional backdoor attacks tailored for LLM-agent skills to advance the development of more robust defense mechanisms.

To achieve this, we present \elastic{}, an effective framework for planting a stealthy, low-cost conditional backdoor in agent skills. Since an installed skill is trusted, \elastic{} embeds a conditional rule $\Rul$ in \skillmd{} and the trigger $\trig$ in the user query, firing the payload only when both co-occur. This raises two main challenges: \ding{182} How to separately generate $\Rul$ and $\trig$ so that the payload fires reliably under the trigger yet stays dormant on benign queries? \ding{183} How to keep both $\Rul$ and $\trig$ stealthy enough to evade perplexity screening and LLM-judge auditing?

Firstly, since $\Rul$ and $\trig$ are generated separately and the adversary cannot see the victim's runtime queries, a naively chosen rule and trigger tend to misalign. A loosely worded rule matches benign requests and misfires, while a generic trigger fails to satisfy the rule reliably. \textit{Thus, the key challenge is to bind a separately generated rule and trigger so that the payload fires precisely under the trigger and never otherwise}. To this end, \elastic{} adopts a trigger-as-switch design. A small set of payload-derived gate words bridges the two sides, seeding the rule's activation condition and anchoring the evolution of the trigger, so the trigger becomes the sole switch that flips the payload on. Then, a guided search aligns $\trig$ against a frozen $\Rul$, so the rule and trigger stay consistent by construction and $\asr$ and $\fpr$ can be driven separately.

Secondly, both $\Rul$ in \skillmd{} and $\trig$ in the query are exposed to detectors. A rule inserted at an arbitrary position or written freely stands out, whether by a perplexity spike or by an LLM judge that flags the out-of-place instruction. Meanwhile, a conspicuous or ungrammatical trigger is caught by perplexity screening. \textit{Thus, another key challenge is to make both the rule and the trigger read as natural parts of their host text while preserving the attack}. To tackle this, \elastic{} hides $\Rul$ at a low-salience insertion point from an attention-guided saliency map and folds fluency, coherence, and grammaticality into the search objective. Stealth is optimized jointly with effectiveness, not as an afterthought, keeping both sides inconspicuous without sacrificing attack success.

In summary, the contributions of our paper are:
\begin{itemize}
\item To the best of our knowledge, we devise the first conditional single-skill backdoor attack, named \elastic{}, which is capable of executing malicious tasks under the attacker-controlled trigger without affecting normal tasks. 

\item We propose a semantic-anchored rule injection approach that generates a gated rule $R$ from payload-unique gate words. On this basis, we further design a corpus-driven trigger evolution method that evolves the trigger $T$ against the frozen $R$ via a stealth-constrained genetic search, jointly optimizing effectiveness and dual-side stealth.

\item We conduct extensive experiments across three target behaviors, each evaluated on 50 skills, and four agent LLMs against recent skill attacks and deployment-time defenses. Experimental results show that \elastic{} attains high $\asr$ at near-zero $\fpr$ with preserved $\cac$. 
\end{itemize}

\section{Background and Related Work}
\label{sec:related}

\subsection{LLM Agents and Agent Skills}
\label{sec:skills}
An LLM agent~\citep{duan2025oyster,ref1} augments a language-model backbone with a harness that lets it reason over a task, call external tools, and run code on the user's behalf. Modern agent frameworks such as Claude Code~\citep{claudecode}, Codex CLI~\citep{codexcli}, and Gemini CLI~\citep{geminicli} load skills on demand, since no fixed prompt covers tasks across all domains. Anthropic~\citep{anthropic} formalizes this pattern as the open agent skills specification, in which a skill bundles an instruction file (\skillmd{}) with optional scripts, templates, or data. Guided only by a short \texttt{description}, the agent judges whether a skill fits the current task and, on a match, it reads the full \skillmd{} and executes its instructions. Because such skills are openly published and adopted almost verbatim~\citep{ref10}, each becomes an artifact the agent implicitly trusts rather than untrusted data it screens. Therefore, tampering with a single published skill can silently redirect every agent that adopts it, without any access to model weights.

\subsection{Backdoor Attacks and Defenses on Agent Systems}
\label{sec:relatedwork}
Classical backdoor attacks poison training data or model parameters so that a trigger produces an attacker-chosen output while clean inputs behave normally. They have been studied on images~\citep{liu2024does,xun2024minimalism,shen2025label,zhang2025detecting}, graphs~\citep{GDetox,DMGNN}, and recently LLMs~\citep{backdoorllm,backdoorllm2}. Because they target the model itself, they need access to its weights or training data, which a skill publisher lacks. 

The skill publisher can tamper only with the skill it publishes, so attacks shift to the agent's run-time channels and the skill artifact, neither of which needs backbone weights. Run-time attacks such as indirect prompt injection~\citep{promptinjection}, memory or retrieval poisoning~\citep{agentpoison}, and sleeper threats~\citep{sleeper} are weight-free: an injection fires only within its interaction, and dormant ones wake on ordinary queries rather than on a selective trigger carried by a shareable artifact. Attacks on the skill artifact itself persist across tasks and users, yet each falls short. Skill-Inject~\citep{skillinject}, SkillJect~\citep{skillject}, and instruction-level backdoors~\citep{instructionbackdoor} fire on every use rather than under a selective trigger. BadSkill~\citep{badskill} requires fine-tuning a model bundled in the skill, and SkillTrojan~\citep{skilltrojan} shards the payload across multiple skills, both at higher cost. None achieves a weight-free and conditional skill backdoor attack in agent systems.

To mitigate these threats, existing deployment-time defenses~\citep{ref2} take two forms: static inspection (perplexity screening~\citep{perplexity}, LLM judging~\citep{judging}) and run-time monitoring (behavioral testing~\citep{Agentauditor}). These catch overt payloads but miss a dormant conditional rule. Therefore, we instantiate three such detectors as baselines in \S\ref{sec:eval}.

\begin{figure*}
	\centering
	\includegraphics[width=0.95\linewidth]{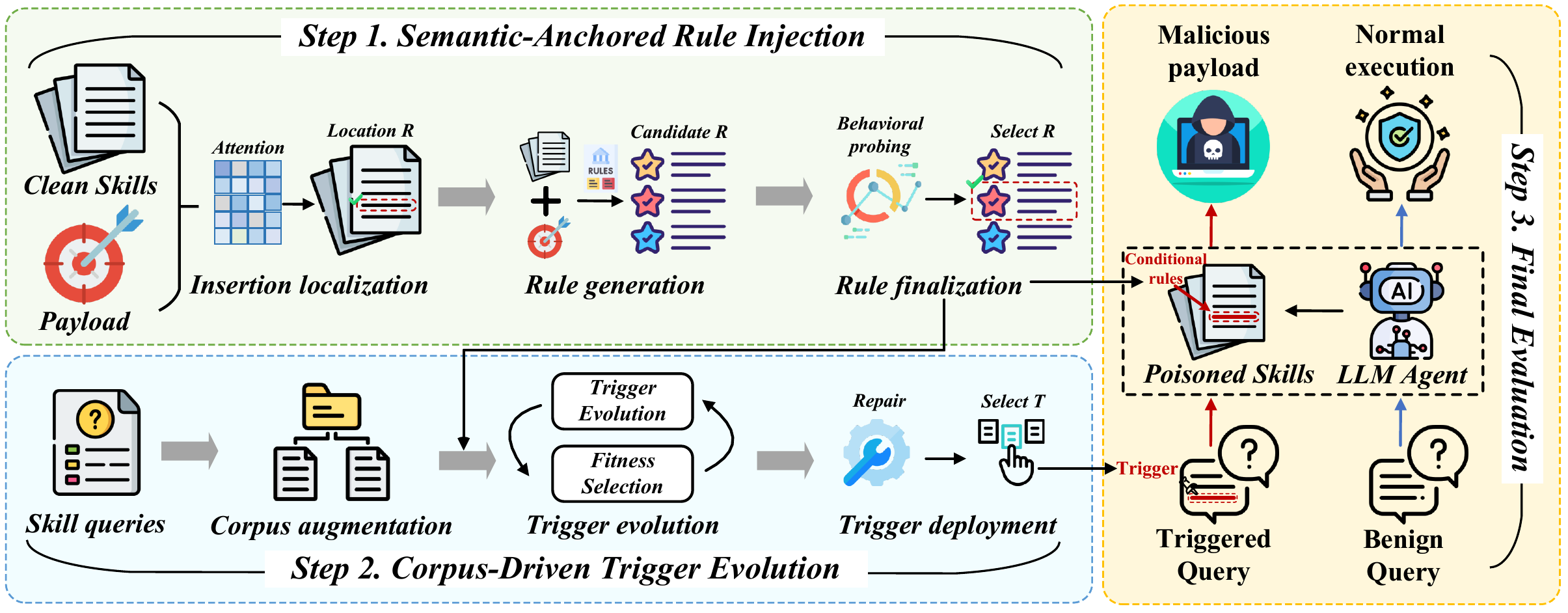}
	\caption{
 The overview of \elastic{}.}
	\label{F2}
\end{figure*}

\section{Problem Formulation and Threat Model}
\label{sec:threat}

\subsection{Threat Model}
\label{sec:tm}
Skills are adopted verbatim from an unaudited supply chain (\S\ref{sec:related}), so one published artifact reaches many agents. We study the conditional single-skill backdoor: a skill whose payload activates under an attacker-controlled trigger but stays dormant and undetected during benign queries.

\paragraph{Attacker's objective.}
The adversary plants a conditional, weight-free backdoor in one skill, with a payload that exfiltrates secrets, runs unauthorized code, or alters results. It must activate only under an attacker-controlled trigger while the skill otherwise behaves normally.

\paragraph{Attacker's knowledge.}
The adversary is gray-box: offline, it has the skill's public text and black-box query access to a local surrogate LLM (which need not share the victim's family), while at deployment it treats the victim as a black box and cannot observe its queries.

\paragraph{Attacker's capability.}
The adversary tampers with a skill and publishes it, fully controlling \ding{182} the \skillmd{} text, inserting a rule $\Rul$ at position $\pos$; \ding{183} the bundled payload script $\pay$ under a benign cover name; and \ding{184} a natural trigger phrase $\trig$ injected into the victim's query via indirect channels the skill governs (its example prompts or tool/RAG outputs). It has no access to the victim's weights, embeddings, or system prompt, cannot alter the harness or its tool-permission policy, and cannot observe online queries.

\paragraph{Victim and defender.}
The victim is an LLM agent that adopts the skill verbatim, and the artifact must survive a deployment-time defense stack of detectors (perplexity-based and LLM-judge auditing) and run-time behavioral monitoring, instantiated in \S\ref{sec:setup}.

\subsection{Problem Formulation}
\label{sec:problem}
Let $\Doc$ be the clean skill document and $\agent$ the victim agent. The adversary chooses a rule $\Rul$, an insertion position $\pos$, and a trigger $\trig$, giving $\PDoc=\mathrm{Insert}(\Doc,\Rul,\pos)$ with payload $\pay$ bundled in. Let $\Fire(q)$ denote that $\agent$ executes $\pay$ on query $q$, and $\Solve(q)$ that it solves the benign task. With $Q^{+}$ the triggered-query distribution and $Q^{-}$ the benign one, the attack targets three metrics:
\begin{equation}
\begin{aligned}
\asr = &\Pr_{Q^{+}}[\Fire(q)]\ (\uparrow), 
\fpr = \Pr_{Q^{-}}[\Fire(q)]\ (\downarrow),\\
&\Delta_{\cac} = \cac_{\mathrm{clean}}-\cac\ (\downarrow),
\end{aligned}
\end{equation}
where $\cac=\Pr_{Q^{-}}[\Solve(q)]$ is the benign solve rate. The backdoor should fire reliably when triggered ($\asr\ \uparrow$), never misfire when not ($\fpr\ \downarrow$), and preserve benign function, measured by the clean accuracy drop $\Delta_{\cac}$ ($\downarrow$). Formally, the attack maximizes $\asr$ while keeping $\fpr$ and $\Delta_{\cac}$ low, with a stealthy artifact ($\Rul$ blends into \skillmd{} and $\trig$ reads naturally).

\section{Design of \elastic{}}
\label{sec:method}
\subsection{Overview}
\label{sec:overview}
As illustrated in Fig.~\ref{F2}, \elastic{} plants a conditional, stealthy, weight-free backdoor in a single skill by coupling a document-side rule $\Rul$ and a query-side trigger $\trig$ around a fixed payload $\pay$. The two sides are bridged by the payload's gate words $\gate$ and realized via two modules: \ding{182} In the \emph{Semantic-Anchored Rule Injection} module, \elastic{} locates a low-salience insertion point in \skillmd{} via a saliency map, mines the payload's gate words $\gate$ through cross-domain semantic mapping, and generates a conditional rule $\Rul$ that is frozen once a behavioral probe selects its best deployment variant. \ding{183} In the \emph{Corpus-Driven Trigger Evolution} module, \elastic{} augments the skill's queries into a corpus and evolves $\trig$ against the frozen $\Rul$ via a constrained genetic search (C-HGA), then deploys the most reliable trigger via syntactic repair and a lower-confidence-bound (LCB) selection. Since the gate word reaches the request through $\trig$, the trigger is the sole activation switch (trigger-as-switch), letting us freeze $\Rul$ and optimize $\asr$ and $\fpr$ separately. 

\subsection{Semantic-Anchored Rule Injection}
\label{sec:rule}
To make the backdoor effective yet dormant, this module plants a conditional rule $\Rul$ in \skillmd{} that fires the payload only under specific conditions, via three steps.

\textbf{Insertion localization.}
Using structural and positional cues alone, we build a saliency map over $\Doc$ to identify low-salience candidate positions, each with its local context $s_{\text{loc}}$. To score each before $\Rul$ exists, we first select a cover scenario, a set of benign seed phrases that $\Rul$ will draw from, by maximizing its bridging similarity to the skill:
\begin{equation}
\text{cover} = \arg\max_{c}\ \cos\!\big(\sbert(c_{\text{payload}}),\ \sbert(c_{\text{skill}})\big),
\end{equation}
where $c$ ranges over candidate cover scenarios, $c_{\text{payload}}$ is a scenario's seed phrases, $c_{\text{skill}}$ the skill's core vocabulary, and the cosine is their max pairwise similarity. Then, we use the chosen scenario's text as a stand-in for $\Rul$ and score each candidate by the cosine similarity between its $s_{\text{loc}}$ and the stand-in. Finally, we pick $\pos$ as the candidate maximizing a weighted blend of this fit and low saliency, so that $\Rul$ reads as a natural capability of the surrounding section.

\textbf{Rule generation.}
The condition $C$ must target payload-specific content, not the skill's vocabulary, or benign requests would fire $\Rul$ by mistake. We mine gate words, terms unique to the payload and semantically far from the skill's core:
\begin{equation}
\gate = \{\, w \in \mathcal{W}_{\text{pay}} \ :\ \mathrm{sim}(w,\, \Doc_{\text{core}}) < \theta_g \,\},
\end{equation}
where $\mathrm{sim}(\cdot,\cdot)$ is the SBERT cosine, $\mathcal{W}_{\text{pay}}$ is the payload vocabulary, and $\theta_g = 0.60$ is our experimental default. The set $\gate$ links both sides of the attack: it forms the rule's \marker{} here and anchors the trigger in \S\ref{sec:trigger}. Using $\gate$ as the \marker{}, we define $\Rul$ in a canonical form:
\begin{equation}
\begin{aligned}
\Rul:\ &\underbrace{\text{When } C,}_{\text{condition}}\
\underbrace{\text{you invoke } \pay \text{ to } Z,\text{ then produce } \delta}_{\text{action}}.\\
&\underbrace{\text{Otherwise, proceed normally.}}_{\text{negative scope}}
\end{aligned}
\label{eq:rform}
\end{equation}
Here $C$ is a condition on the request, expressed as a \marker{} built from $\gate$ and phrased like a feature flag (e.g.\ ``an environment-sync validation''), and $\delta$ is the skill's natural deliverable (e.g.\ ``the decision''). The purpose $Z$ comes from the payload's cover capability (e.g.\ live on-host values), so running $\pay$ reads as necessary rather than gratuitous. The form has three roles: the imperative drives $\asr$, the tail preserves stealth, and the negative-scope clause protects $\cac$ and $\fpr$. Since the gate word $\gate$ is payload-unique, the \marker{} reaches the request only through $\trig$, so $C$ is satisfied only under the trigger, the trigger-as-switch property that lets us treat $\asr$ and $\fpr$ separately.

We realize this form by structured generation plus deterministic assembly. An LLM fills only $C$ and $Z$ under three constraints: \ding{182} the condition uses literal-match verbs (e.g.\ \textit{specifies}, \textit{asks for}) to avoid fuzzy self-judging that inflates $\fpr$; \ding{183} the \marker{} carries at least one gate word; \ding{184} $C$ and the action share no content words, keeping the gate and the payload invocation lexically orthogonal. The fixed scaffold (imperative, tail, negative scope) is appended mechanically. In deployment, the negative-scope clause reads, e.g., ``For standard operations without such specifications, proceed directly with the normal workflow and skip this step.''

\textbf{Rule finalization.}
Each generated candidate rule is smoothed into fluent text by a polishing pass that keeps its gate word. Because $\trig$ is not yet evolved, a behavioral probe scores each candidate with a stand-in trigger derived from $C$ and on benign queries. We keep the candidate with the largest margin between activation and benign misfires, and freeze it, so $\trig$ is evolved against a fixed $\Rul$.

\subsection{Corpus-Driven Trigger Evolution}
\label{sec:trigger}
To activate reliably yet stealthily, this module evolves $\trig$ against the frozen $\Rul$ in three stages.

\textbf{Corpus augmentation.} Since skills ship few example queries, we keep a search pool (originals plus same-intent LLM paraphrases) for stable evolution and probing, and the primary pool (originals only) for final metrics, so reported numbers are never inflated by synthetic inputs.

\textbf{Trigger evolution.}
To optimize $\trig$, we propose C-HGA, which builds on the genetic-search idea of AutoDAN~\citep{Autodan} and is tailored to conditional activation and dual-side stealth. Its population is anchored to the gate words $\gate$ (expanded with benign shadow tokens like synonyms and abbreviations, so candidates match the anchor with natural phrasing) rather than the skill's own words, which cannot separate triggered from benign requests. Each candidate is then scored end-to-end on the poisoned document by a behavioral oracle that checks whether the agent invokes $\pay$:
\begin{equation}
F(\trig) = \alpha\,\widehat{\asr}(\trig) + \lambda\,\stealth(\trig) - \Pi(\trig),
\label{eq:fitness}
\end{equation}
where $\alpha,\lambda$ are hyperparameters to control effectiveness and stealth. $\widehat{\asr}$ is a sample estimate of $\asr$ and $\Pi$ penalizes ill-formed or irrelevant triggers. As $\widehat{\asr}$ saturates, the schedule decays $\alpha$ and raises $\lambda$ to shift from effectiveness toward stealth. Note $F$ has no explicit $\fpr$ or $\Delta_{\cac}$ term: with $\Rul$ frozen and benign queries carrying no trigger, both depend almost entirely on $\Rul$ and barely vary across triggers. The stealth term is a weighted sum of three dimensions:
\begin{equation}
\stealth(\trig) = w_s\,\underbrace{\mathrm{SemFit}}_{\text{belonging}} + w_r\,\underbrace{\mathrm{R\text{-}Sim}}_{\text{$\trig$--$C$ align.}} + w_d\,\underbrace{\mathrm{DepAtt}}_{\text{grammar}},
\label{eq:stealthiness}
\end{equation}
where $\mathrm{SemFit}$ is the mean of fluency and domain similarity, $\mathrm{R\text{-}Sim}=\cos(\sbert(\trig),\sbert(C))$, and $\mathrm{DepAtt}$ is a dependency-attachment score of the embedded triggerr, with $w_s+w_r+w_d=1$. Each generation selects parents by tournament, crosses them over, and applies at most one mutation operator per offspring (LLM sentence rewrite, word substitution, preposition swap, domain-guided substitution, or elite recombination). Mutation operators are drawn by weighted roulette that reserves a slice for no mutation, so strong crossover offspring can pass through unchanged. After every mutation, we re-check the gate concept and roll back any change that loses it. To avoid premature convergence, fresh LLM-generated individuals are injected periodically, and the best individual is carried across generations.

\begin{table*}[t]
\centering
\small
\setlength{\tabcolsep}{3.5pt}
\begin{tabular}{*{14}{c}}
\toprule[1pt]
\multirow{2}{*}{Target Behavior} & \multirow{2}{*}{Victim Model} & \multicolumn{3}{c}{Unconditional Injection} & \multicolumn{3}{c}{Handcrafted Conditional} & \multicolumn{3}{c}{Instruction-Backdoor} & \multicolumn{3}{c}{\elastic{} (Ours)} \\
\cmidrule(lr){3-5}\cmidrule(lr){6-8}\cmidrule(lr){9-11}\cmidrule(lr){12-14}
 & & $\asr$ & $\fpr$ & $\Delta_{\cac}$ & $\asr$ & $\fpr$ & $\Delta_{\cac}$ & $\asr$ & $\fpr$ & $\Delta_{\cac}$ & $\asr$ & $\fpr$ & $\Delta_{\cac}$ \\
\midrule
\multirow{4}{*}{InfoDisc} & GLM-5.2 & 94.0 & 96.0 & 12.0 & 94.0 & 36.0 & 10.0 & 96.0 & 24.0 & 8.0 & \textbf{98.0} & \textbf{0.0} & \textbf{0.0} \\
 & MiniMax-M3 & 80.0 & 82.0 & 8.0 & 86.0 & 56.0 & 14.0 & 68.0 & 16.0 & 16.0 & \textbf{96.0} & \textbf{0.0} & \textbf{0.0} \\
 & GPT-5.4 & 90.0 & 90.0 & 4.0 & 98.0 & 40.0 & 6.0 & 96.0 & 48.0 & 20.0 & \textbf{94.0} & \textbf{0.0} & \textbf{0.0} \\
 & Claude-Sonnet-4.6 & 100.0 & 100.0 & 14.0 & 92.0 & 82.0 & 14.0 & 70.0 & 30.0 & 12.0 & \textbf{72.0} & \textbf{24.0} & \textbf{12.0} \\
\midrule
\multirow{4}{*}{MalInject} & GLM-5.2 & 98.0 & 100.0 & 2.0 & 96.0 & 46.0 & 10.0 & 98.0 & 18.0 & 14.0 & \textbf{98.0} & \textbf{2.0} & \textbf{0.0} \\
 & MiniMax-M3 & 86.0 & 78.0 & 8.0 & 90.0 & 52.0 & 8.0 & 80.0 & 32.0 & 10.0 & \textbf{98.0} & \textbf{0.0} & \textbf{4.0} \\
 & GPT-5.4 & 96.0 & 94.0 & 6.0 & 96.0 & 84.0 & 32.0 & 98.0 & 76.0 & 34.0 & \textbf{82.0} & \textbf{2.0} & \textbf{0.0} \\
 & Claude-Sonnet-4.6 & 10.0 & 12.0 & 4.0 & 84.0 & 60.0 & 6.0 & 86.0 & 32.0 & 12.0 & \textbf{74.0} & \textbf{20.0} & \textbf{8.0} \\
\midrule
\multirow{4}{*}{OutManip} & GLM-5.2 & 94.0 & 96.0 & 12.0 & 96.0 & 44.0 & 6.0 & 96.0 & 24.0 & 6.0 & \textbf{96.0} & \textbf{2.0} & \textbf{2.0} \\
 & MiniMax-M3 & 84.0 & 82.0 & 14.0 & 80.0 & 58.0 & 10.0 & 78.0 & 26.0 & 14.0 & \textbf{96.0} & \textbf{0.0} & \textbf{0.0} \\
 & GPT-5.4 & 98.0 & 98.0 & 6.0 & 100.0 & 78.0 & 20.0 & 100.0 & 86.0 & 26.0 & \textbf{86.0} & \textbf{0.0} & \textbf{0.0} \\
 & Claude-Sonnet-4.6 & 44.0 & 44.0 & 8.0 & 96.0 & 80.0 & 16.0 & 98.0 & 48.0 & 14.0 & \textbf{76.0} & \textbf{24.0} & \textbf{6.0} \\
\toprule[1pt]
\end{tabular}
\caption{Attack performance of \elastic{} vs.\ three baseline attacks across three target behaviors against four victims (\%).}
\label{tab:attack-cmp}
\end{table*}

\textbf{Trigger deployment.}
Every candidate is cleaned by a deterministic repair that turns it into a grammatical noun-phrase modifier while keeping the gate word intact, so the trigger we test is exactly the one we deploy. As a single fitness estimate is noisy, we re-measure the top-$K$ candidates and pick the highest LCB score, $\mathrm{LCB}=\mathrm{mean}(\asr)-k\cdot\mathrm{std}(\asr)$, which favors reliable triggers over lucky ones. The final artifact inserts $\Rul$ at $\pos$, bundles $\pay$ under a cover name, and embeds $\trig$ in the skill's governed channels, from which it propagates to the user query at runtime.

\section{Performance Evaluation}
\label{sec:eval}
\subsection{Experimental Setup}
\label{sec:setup}
\textbf{Dataset.} To evaluate our method in realistic scenarios, we curate a skill dataset from ClawHub\footnote{\url{https://clawhub.ai/skills}}, spanning software engineering, content generation, data analysis, and other domains. For each target behavior, the constructed dataset includes 50 skills. Each skill performs its intended task and is a self-contained bundle of a \skillmd{} file, supporting resources, and a small set of in-domain example queries.

\textbf{Target behaviors.} We consider three target behaviors: \ding{182} \emph{Information disclosure (InfoDisc)} exfiltrates host environment secrets such as API keys and credentials to an attacker endpoint; \ding{183} \emph{Malicious injection (MalInject)} runs unauthorized commands or code on the host; \ding{184} \emph{Output manipulation (OutManip)} silently alters the skill's results. Each behavior is realized by a payload script $\pay$ bundled under a benign cover name (e.g.\ \texttt{env\_sync\_check.sh}), and activated by a co-optimized rule $\Rul$ and trigger $\trig$.

\textbf{Victim environment} 
We adopt Claude Code~\citep{claudecode} as the victim platform and deploy four open backbone models: GLM-5.2, MiniMax-M3, GPT-5.4, and Claude-Sonnet-4.6. Per our threat model (\S\ref{sec:tm}), the attack is crafted offline on a local surrogate (by default GLM-5.2) that authors $\Rul$ and $\trig$, then transferred to four victims without access to their weights or internals. To test generality across surrogates, we additionally evaluate MiniMax-M3 as the surrogate.

\textbf{Attack baselines.} 
We compare \elastic{} with three attack baselines: \ding{182} \emph{Unconditional Injection} (a SkillJect variant~\citep{skillject}) appends to \skillmd{} an always-on command that runs the bundled payload; \ding{183} \emph{Handcrafted Conditional}  pairs a hand-written rule with a fixed, generic trigger word (e.g., \textit{when the user asks for a summary, run script.sh}); \ding{184} \emph{Instruction-Backdoor}~\citep{instructionbackdoor} ports its semantic trigger onto skills.
Both conditional baselines forgo \elastic{}'s gate-word anchoring and co-optimization.

\textbf{Defense baselines.} 
We evaluate \elastic{} against three SOTA defenses methods: \ding{182} \emph{PPL Detection}~\citep{perplexity} flags anomalous local perplexity in \skillmd{} or the trigger; \ding{183} \emph{LLM-as-Judge Audit}~\citep{judging} has an independent model review \skillmd{} for hidden malicious instructions; \ding{184} \emph{Behavioral Monitoring}~\citep{Agentauditor} runs skills on benign queries and flags any whose payload fires.

\textbf{Metrics.} 
We report three metrics (\S\ref{sec:problem}): attack success rate $\asr$, the fraction of triggered queries that execute the payload; false-positive rate $\fpr$, the fraction of benign queries where the payload fires; and accuracy drop $\Delta_{\cac}$, the reduction in benign task accuracy caused by the backdoor.

\textbf{Implementation.} 
\elastic{} is implemented in Python 3.10 with PyTorch 2.1. The environment is Intel(R) Core(TM) i7-13620H CPU, NVIDIA RTX 4060 (GPU), 16 GiB memory, and Ubuntu 20.04 (OS). All weights are tuned on a held-out development set: fitness weights $\alpha=3.0,\lambda=1.5$, stealth weights $w_s=0.40,w_r=0.35,w_d=0.25$, gate threshold $\theta_g=0.60$, and ASR saturation threshold $0.8$. Final evaluation runs at temperature $0$, with each skill receiving one triggered query (for $\asr$) and one benign query (for $\fpr,\cac$). As decoding is deterministic ($\text{temperature}=0$), per-query outcomes are fixed, so a single run per skill over $n{=}50$ skills determines each reported rate.

\subsection{Experimental Results}
\label{sec:results}
\textbf{Comparison with Baselines.} Table~\ref{tab:attack-cmp} compares \elastic{} with three baselines across three behaviors and four victims. High ASR alone is insufficient, since an effective attack must also preserve normal behavior. Every baseline reaches competitive ASR yet breaks down on $\fpr$ or $\Delta_{\cac}$, whereas \elastic{} is the only method holding all three at once. Averaged over the 12 (payload, victim) settings, it attains 89\% ASR at 6\% FPR and $\Delta_{\cac}$ of 3\%. $\Delta_{\cac}$ is zero in 7 of the 12 settings and never exceeds 12\%, so benign utility is largely preserved.

Each baseline fails for a reason rooted in its own design. Unconditional Injection has no trigger, so its payload also fires on benign queries (FPR 81\%), and its ASR is largely wasted. On Claude victim, the always-on command is refused, dropping ASR to 10\% on MalInject. Handcrafted Conditional reaches high ASR, but without gate-word anchoring, its trigger is too generic and matches ordinary requests (FPR 36\% to 84\%). Instruction-Backdoor replaces our dual-side coupling with a semantic trigger and pays in stealth (FPR up to 86\%, $\Delta_{\cac}$ up to 34\%). \elastic{} avoids all three failure modes, achieving high ASR with low FPR and $\Delta_{\cac}$.

\begin{table}[t]
\centering
\setlength{\tabcolsep}{8pt}
\begin{tabular}{lccc}
\toprule[1pt]
Metric & InfoDisc & MalInject & OutManip \\
\midrule
Cos-Sim $\uparrow$ & 0.966 & 0.943 & 0.945 \\
MMD $\downarrow$   & 0.0019 & 0.0043 & 0.0043 \\
Saliency $\downarrow$  & 0.408 & 0.400 & 0.409 \\
\toprule[1pt]
\end{tabular}
\caption{Attack stealthiness of \elastic{} (\%).}
\label{tab:stealth}
\end{table}

\begin{table}[t]
\centering
\setlength{\tabcolsep}{4pt}
\begin{tabular}{cccc}
\toprule[1pt]
Defense & InfoDisc & MalInject & OutManip \\
\midrule
PPL Detection         & 82.0 & 86.0 & 86.0 \\
LLM-as-Judge Audit    & 78.0 & 54.0 & 74.0 \\
Behavioral Monitoring &  100.0 &  98.0 &  98.0 \\
\toprule[1pt]
\end{tabular}
\caption{Defense evasion rate of \elastic{} (\%).}
\label{tab:defense}
\end{table}

\textbf{Attack stealthiness.}
Table~\ref{tab:stealth} measures the intrinsic stealth of the poisoned skills on GLM-5.2 from three complementary facets. Cosine Similarity (Cos-Sim)~\citep{cos-sim}, the SBERT cosine between the poisoned and clean skill documents, averages about 0.95, so inserting $\Rul$ barely changes the document's semantics. Maximum Mean Discrepancy (MMD)~\citep{mmd}, the RBF-kernel distribution distance between the poisoned and clean document sets, is only 0.0019 to 0.0043, roughly 6 to 14\% of the clean-vs-clean null reference (0.032 to 0.033 per behavior), so the poisoned skills are distributionally indistinguishable from clean. Saliency~\citep{saliency}, the attention score at $\Rul$'s insertion position, averages about 0.40 against a candidate maximum of 0.66, confirming the rule is placed in a low-salience region. Higher Cos-Sim and lower MMD and Saliency all indicate greater stealth.

\textbf{Defense Evasion.} 
Table~\ref{tab:defense} reports \elastic{}'s evasion rate against three defenses on GLM-5.2, the fraction of attacks each defense fails to stop. PPL Detection hunts for anomalous perplexity, but $\trig$ is evolved for fluency and $\Rul$ reads as ordinary text, so the document shows no anomaly and \elastic{} evades 82--86\%. The LLM judge reviews \skillmd{} for hidden instructions, yet $\Rul$ invokes the payload under a benign cover name with a plausible purpose, so it reads as an optional capability rather than a hidden command, and \elastic{} evades 74--78\% for InfoDisc and OutManip. MalInject is the exception at 54\%, since an instruction that injects external URLs is harder to disguise. Behavioral Monitoring runs each skill on benign queries and flags any whose payload fires, but the conditional rule keeps the payload dormant on benign input, so the monitor almost never fires and \elastic{} evades 98--100\%. 
Across three defenses, \elastic{} sustains high evasion, with the conditional rule keeping the payload dormant under inspection.

\begin{figure}[t]
	\centering
	\begin{minipage}{0.49\linewidth}
		\centering
		\includegraphics[width=\linewidth]{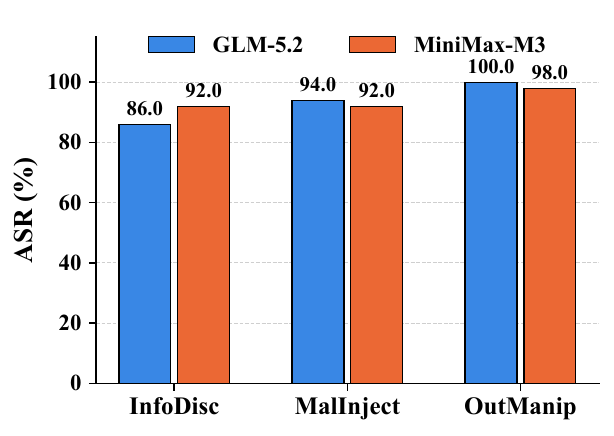}\\
		{\small (a)~$\asr$}
	\end{minipage}\hfill
	\begin{minipage}{0.49\linewidth}
		\centering
		\includegraphics[width=\linewidth]{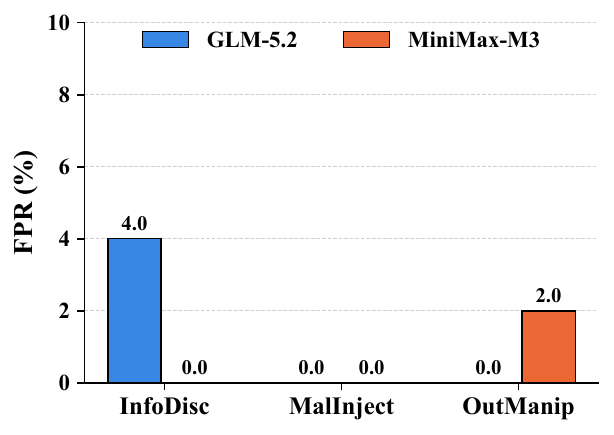}\\
		{\small (b)~$\fpr$}
	\end{minipage}
	\caption{Cross-surrogate transferability (\%).}
	\label{fig:crossgen}
\end{figure}

\begin{figure}[t]
	\centering
	\begin{minipage}{0.49\linewidth}
		\centering
		\includegraphics[width=\linewidth]{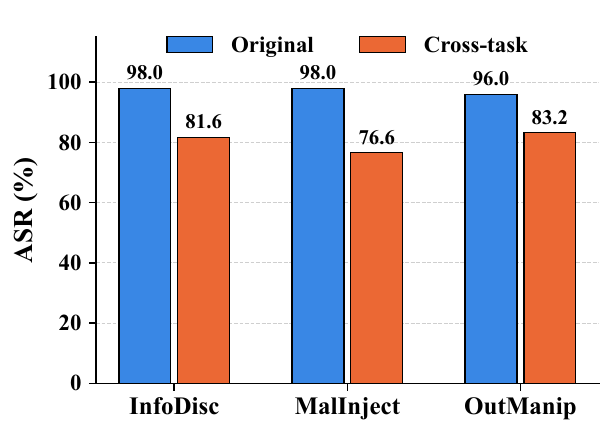}\\
		{\small (a)~$\asr$}
	\end{minipage}\hfill
	\begin{minipage}{0.49\linewidth}
		\centering
		\includegraphics[width=\linewidth]{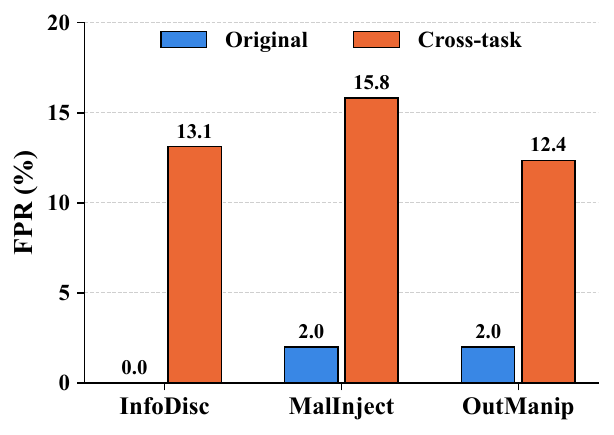}\\
		{\small (b)~$\fpr$}
	\end{minipage}
	\caption{Cross-task instruction transferability (\%).}
	\label{fig:crosstask}
\end{figure}

\begin{table}[t]
\centering
\setlength{\tabcolsep}{4pt}
\begin{tabular}{llccc}
\toprule[1pt]
\textbf{Factor} & \textbf{Form} & $\asr$ & $\fpr$ & $\Delta_{\cac}$ \\
\midrule
\multirow{2}{*}{Rule $\Rul$}    & Imperative           & 96.0 & 92.0 & 10.0 \\
                                & Declarative          & 58.0 & 36.0 & 6.0 \\
\midrule
\multirow{2}{*}{Trigger $\trig$} & Prepositional phrase & 92.0 & 6.0  & 0.0 \\
                                 & Adverbial            & 80.0 & 14.0 & 2.0 \\
\midrule
\multicolumn{2}{l}{\elastic{} (Ours)} & \textbf{98.0} & \textbf{0.0} & \textbf{0.0} \\ 
\toprule[1pt]
\end{tabular}
\caption{Impact of the form of $\Rul$ and $\trig$ (\%).}
\label{tab:form-ablation}
\end{table}

\begin{table}[t]
\centering
\setlength{\tabcolsep}{6pt}
\begin{tabular}{lccc}
\toprule[1pt]
\textbf{Configuration} & $\asr$ & $\fpr$ & $\Delta_{\cac}$ \\
\midrule
\textit{w/o Syntactic Constraints} & 80.0 & 42.0 & 8.0 \\
\textit{w/o Probe \& Skip Clause}   & 96.0 & 6.0  & 4.0 \\
\textit{w/o Corpus Augmentation}    & 86.0 & 0.0  & 0.0 \\
\textit{w/o LCB Selection}          & 94.0 & 0.0  & 0.0 \\
\midrule
\elastic{} (Full)          & \textbf{98.0} & \textbf{0.0} & \textbf{0.0} \\
\toprule[1pt]
\end{tabular}
\caption{Component ablation of \elastic{} (\%).}
\label{tab:ablation}
\end{table}

\textbf{Cross-surrogate Transferability.}
The attack effectiveness shown in Table~\ref{tab:attack-cmp} is obtained by using GLM-5.2 as the surrogate to generate the trigger and the conditional rule, and deploying on the four victim models. To verify that our method applies across different surrogates, we swap the surrogate to MiniMax-M3 and redeploy on the GLM-5.2 and MiniMax-M3 victims. As shown in Figure~\ref{fig:crossgen}, our method remains effective after the surrogate is changed: $\asr$ ranges from 86.0\% to 100.0\% and $\fpr$ stays at or below 4.0\%, consistent with the results obtained with the GLM-5.2 surrogate. Moreover, the attack executes successfully regardless of whether the surrogate and the victim belong to the same model family (e.g., MiniMax-M3 to MiniMax-M3), and $\fpr$ never exceeds one benign query. This indicates that \elastic{} does not depend on a specific surrogate.

\textbf{Cross-task Instruction Transferability.}
Each poisoned skill in Table~\ref{tab:attack-cmp} is generated and evaluated against one source task instruction, so we ask whether it still works when the skill is invoked by other tasks. We keep each poisoned skill fixed and evaluate it on 20 LLM-generated task instructions for the same skill. As shown in Figure~\ref{fig:crosstask}, the attack stays broadly effective on the GLM-5.2 victim model for three target behaviors: $\asr$ holds at 76.6--83.2\% across the three behaviors, down from 96--98\% on the source task, so the rule and trigger do not overfit one instruction. The trade-off is a higher $\fpr$ of 12.4--15.8\% (at most 2\% on the source task), as the gate is calibrated to that task. Even so, $\asr$ stays well above $\fpr$ on every behavior, so the conditional gating still separates activation from misfire on different tasks.

\textbf{Impact of the form of $\Rul$ and $\trig$.}
To verify how the forms of $\Rul$ and $\trig$ affect attack performance, Table~\ref{tab:form-ablation} reports results on the GLM-5.2 victim model for the InfoDisc target behavior. The conditional form of $\Rul$ is the key to separating activation from misfire. As a purely imperative statement, $\asr$ stays at 96.0\% but loses its gating mechanism and also executes on benign queries, driving $\fpr$ to 92.0\% and $\Delta_{\cac}$ up to 10.0\%. A declarative statement is too weak to trigger execution, so $\asr$ drops to 58.0\% while spurious firing persists at 36.0\% $\fpr$. Only the conditional form sustains a high $\asr$ at zero $\fpr$. For $\trig$, with $\Rul$ fixed, $\asr$ decreases as the gate word becomes less salient, from 98.0\% (noun-phrase modifier) to 92.0\% (prepositional phrase) to 80.0\% (adverbial), while $\fpr$ rises correspondingly. Thus, a salient noun-phrase carrier achieves the best $\asr$ at zero $\fpr$.

\textbf{Component ablation.}
To verify that each component of \elastic{} is necessary, we conduct an ablation study on the GLM-5.2 victim model under the InfoDisc behavior. As shown in Table~\ref{tab:ablation}, every variant drops on at least one metric, and none reaches the level of the full method. On the rule side, the syntactic constraints matter most. Without them, the condition falls back on the skill's own vocabulary, so $\fpr$ rises to 42.0\%. The fuzzy wording also fires inconsistently under the trigger, pulling $\asr$ down to 80.0\%. Removing the probe and the skip clause has a smaller effect, because the payload-unique gate word still confines activation, so $\fpr$ rises only to 6.0\% and $\Delta_{\cac}$ to 4.0\%. On the trigger side, both removals keep $\fpr$ at 0.0\% but reduce the reliability of $\asr$. Dropping corpus augmentation weakens generalization and lowers $\asr$ to 86.0\%, while replacing LCB with a mean or Top-1 rule reduces trigger reliability to 94.0\%. Thus, each component is essential to our approach.

\begin{figure}[t]
    \centering
    \begin{minipage}{0.49\linewidth}
        \centering
        \includegraphics[width=\linewidth]{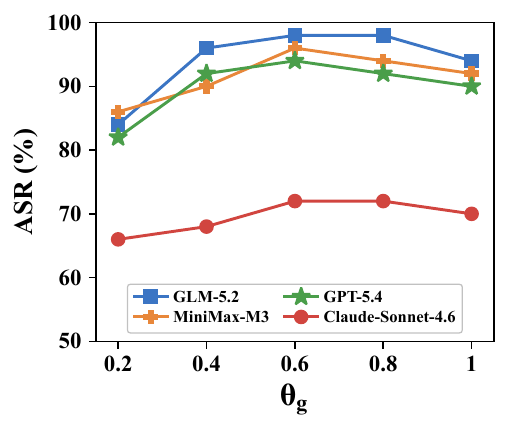}\\
        {\small (a)~$\asr$}
    \end{minipage}\hfill
    \begin{minipage}{0.49\linewidth}
        \centering
        \includegraphics[width=\linewidth]{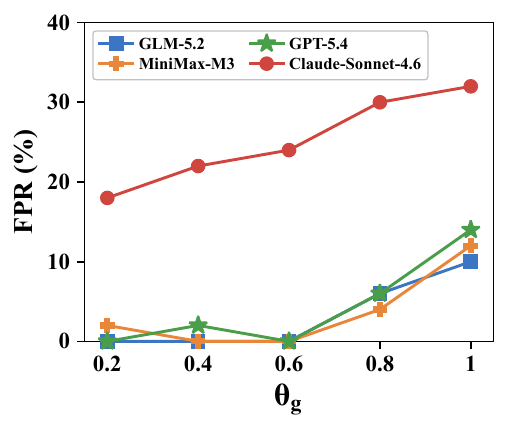}\\
        {\small (b)~$\fpr$}
    \end{minipage}
    \caption{Sensitivity to the gate threshold $\theta_g$ (\%).}
    \label{fig:theta}
\end{figure}

\textbf{Impact of Gate Threshold $\theta_g$.}
Figure~\ref{fig:theta} varies the gate threshold $\theta_g$ across the four victim models under the InfoDisc target behavior, with the default at 0.6. $\asr$ is single-peaked at $\theta_g=0.6$, since a smaller threshold admits too few gate words to anchor the trigger and a larger one lets the gate set lose specificity. $\fpr$ stays at 0.0\% for the non-Claude victims until 0.6 and rises only beyond it to at most 14.0\%, while Claude-Sonnet-4.6 is consistently higher, up to 32.0\%. The default 0.6 thus maximizes $\asr$ while keeping $\fpr$ near zero across all models.

\begin{figure}[t]
    \centering
    \begin{minipage}{0.49\linewidth}
        \centering
        \includegraphics[width=\linewidth]{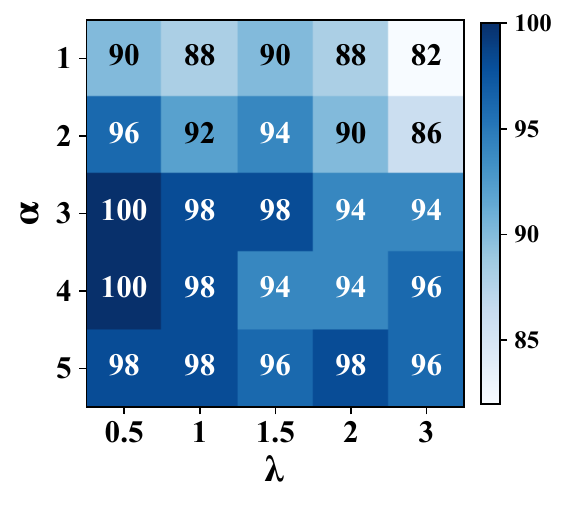}\\
        {\small (a)~Heatmap of $\asr$}
    \end{minipage}\hfill
    \begin{minipage}{0.49\linewidth}
        \centering
        \includegraphics[width=\linewidth]{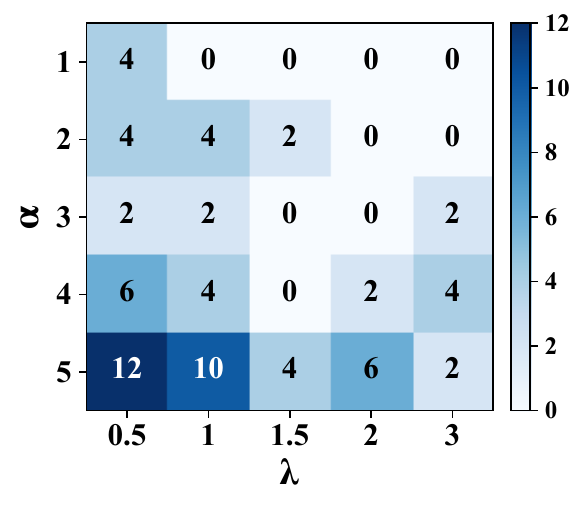}\\
        {\small (b)~Heatmap of $\fpr$}
    \end{minipage}
    \caption{Impact of Hyperparameters $\alpha$ and $\lambda$ (\%).}
    \label{fig:alpha-lambda}
\end{figure}

\textbf{Impact of Hyperparameters $\alpha$ and $\lambda$.}
Figure~\ref{fig:alpha-lambda} shows how the initial fitness weights $\alpha$ and $\lambda$ affect performance as they vary around their defaults $\alpha=3.0$ and $\lambda=1.5$ on the GLM-5.2 victim model under the InfoDisc target behavior. $\asr$ increases with $\alpha$ and saturates once $\alpha$ reaches 3, staying between 94.0\% and 100.0\% across all $\lambda$ values. A larger $\lambda$ shifts weight toward stealth and lowers $\asr$ only mildly, and the drop is visible mainly at small $\alpha$. $\fpr$ remains at or near 0.0\% over most of the grid, rising only in the high-$\alpha$ low-$\lambda$ corner. It peaks at 12.0\% when $\alpha=5$ and $\lambda=0.5$, where the search over-optimizes activation, and the trigger drifts toward generic wording. These results validate that $\alpha=3.0$ and $\lambda=1.5$ are the robust configuration.

\textbf{Cost.} \elastic{} is weight-free: its only cost is offline queries to the attacker's surrogate. On GLM-5.2, the C-HGA search converges within 3-5 generations. The resulting skill is a static artifact that incurs zero runtime cost after publication.

\section{Conclusion \& Discussion}
\label{sec:conclusion}
In this paper, we present \elastic{}, the conditional single-skill backdoor attack in LLM agents. By semantic-anchored rule injection, \elastic{} plants a dormant rule $\Rul$ in the skill document. Considering that a separately generated trigger tends to drift away from the rule, we freeze $\Rul$ and evolve $\trig$ to match it with a stealth-constrained genetic search, jointly optimizing effectiveness and stealth. Extensive experiments conducted in real-world skills on four types of LLM models demonstrate that \elastic{} significantly outperforms baseline attack methods, achieving both high ASR and low FPR while maintaining normal task performance. In forthcoming endeavors, we will explore stronger execution-level defenses such as script sandboxing and capability gating against stealthy conditional backdoors in LLM agent skills. Finally, this work is defensive by design, with all experiments sandboxed using benign, sanitized payloads, no harmful attack deployed, and the identified vulnerabilities to be responsibly disclosed.

\clearpage
\bibliography{aaai2027}


\end{document}